\pdfoutput=1

\documentclass[11pt]{article}

\usepackage[]{acl}

\usepackage{times}
\usepackage{latexsym}
\usepackage{times}
\usepackage{latexsym}
\usepackage{graphicx}
\usepackage{adjustbox}
\usepackage{booktabs}
\usepackage{todonotes}
\usepackage{multirow}
\usepackage[ruled,vlined]{algorithm2e}

\usepackage[T1]{fontenc}

\usepackage[utf8]{inputenc}

\usepackage{microtype}

\title{Extending the Scope of Out-of-Domain: Examining QA models in multiple subdomains}

\author{
  \textbf{Chenyang Lyu}$^\dag$
  ~~~~ \textbf{Jennifer Foster}$^\dag$~~~~ \textbf{Yvette Graham}$^\P$ \\
  $^\dag$ School of Computing, Dublin City University, Dublin, Ireland \\
  $^\P$ School of Computer Science and Statistics, Trinity College Dublin, Dublin, Ireland\\
  \texttt{chenyang.lyu2@mail.dcu.ie}, \texttt{jennifer.foster@dcu.ie}, \texttt{ygraham@tcd.ie} \\
  }

\begin{document}
\maketitle

\begin{abstract}

Past works that investigate out-of-domain performance of QA systems have mainly focused on \textit{general domains} (e.g. news domain, wikipedia domain), underestimating the importance of \textit{subdomains} defined by the internal characteristics of QA datasets. In this paper, we extend the scope of ``out-of-domain'' by splitting QA examples into different subdomains according to their several internal characteristics including \textit{question type, text length, answer position}. We then examine the performance of QA systems trained on the data from different subdomains. Experimental results show that the performance of QA systems can be significantly reduced when the train data and test data come from different subdomains. These results question the generalizability of current QA systems in multiple subdomains, suggesting the need to combat the bias introduced by the internal characteristics of QA datasets.

\end{abstract}

\section{Introduction}
Examining the out-of-domain performance of QA systems is an
important focus of the research community due to its direct connection to the generalizability and robustness of QA systems especially in production environments~\cite{jia-liang-2017-adversarial, chen-etal-2017-reading,talmor-berant-2019-multiqa, fisch2019mrqa, shakeri-etal-2020-end-wangzhiguo, lyu2021improving}, 
Even though previous studies mostly focus on coarse-grained \textit{general domains}~\cite{ruder-sil-2021-multi-domain}, the
 importance of finer-grained \textit{subdomains} defined by the internal characteristics of QA datasets cannot be neglected. For example, several studies exploring specific internal characteristics of QA datasets have been carried out, including \newcite{ko-etal-2020-look-first-sent}, who reveal that the sentence-level answer position is a source of bias for QA models, and  \newcite{sen-saffari-2020-models-learn-from-qa-overlap} who investigate the effect of word-level question-context overlap.

\begin{figure}
    \centering
    \includegraphics[width=\linewidth]{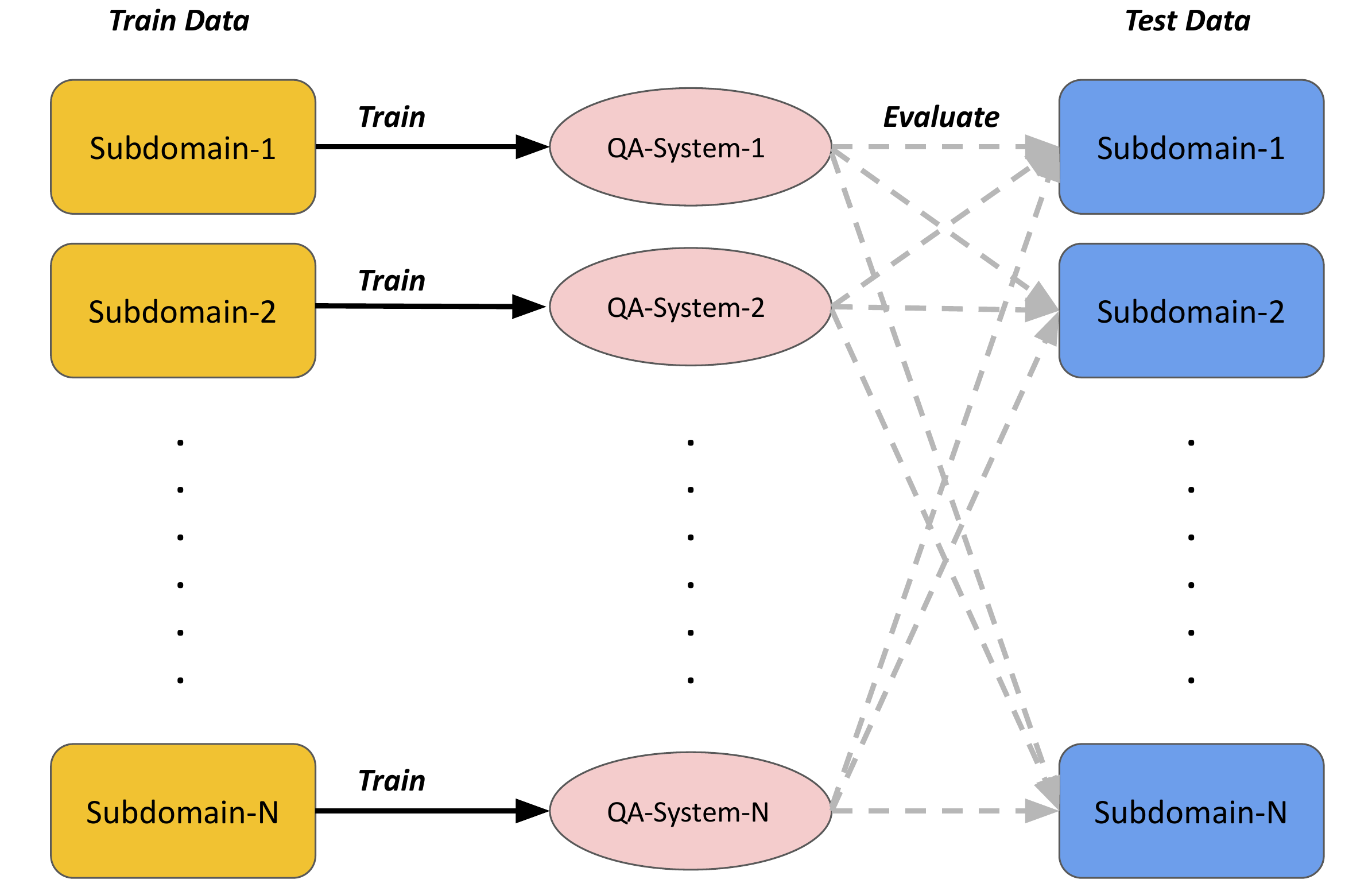}
    \caption{We train QA systems on each subdomain and evaluate each system on all subdomains}
    \label{fig:idea_sketch}
\end{figure}
Building on this prior work as well
as the definition and discussion of \textit{subdomain} in \newcite{DBLP:conf/lrec/PlankS08, DBLP:conf/konvens/Plank16, varis-bojar-2021-sequence}, we extend the scope of out-of-domain with a view to assessing the generalizability and robustness of QA systems by investigating their \textit{out-of-subdomain} performance.
As shown in Figure~\ref{fig:idea_sketch}, we split the QA dataset into different \textit{subdomains} based on its internal characteristics. Then we use the QA examples in each subdomain to train corresponding QA systems and evaluate their performance on all subdomains.

We focus on extractive QA (EQA) as it is not only an important task in  itself~\cite{zhang2020machine} but also the crucial \textit{reader} component in the retriever-reader model for Open-domain QA~\cite{chen-etal-2017-reading,chen-yih-2020-open}.
In experiments with SQuAD 1.1~\cite{rajpurkar-etal-2016-squad1.1} and NewsQA~\cite{trischler-etal-2017-newsqa}, we split the 
data into subdomains based on \textit{question type, text length (context, question and answer)} and \textit{answer position}. We then train QA systems on each subdomain and examine their performance on each subdomain.  Results show that QA systems tend to perform worse when train and test data come from different subdomains, particularly 
those defined by \textit{question type, answer length} and \textit{answer position}.

\section{Experiments}

We employ the QA datasets SQuAD1.1~\cite{rajpurkar-etal-2016-squad1.1} and NewsQA~\cite{trischler-etal-2017-newsqa}. For SQuAD1.1 we use the official dataset released by \newcite{rajpurkar-etal-2016-squad1.1} and for NewsQA we use the data from MRQA~\cite{fisch2019mrqa}. For question classification, we use the dataset from \newcite{li-roth-2002-learning_question_classification}. We use the BERT-base-uncased model from Huggingface~\cite{Wolf2019HuggingFacesTS} for both question classification and QA.

We adopt the following setup for training and evaluation: We split the original training set $D$ into several disjoint subdomains $D_{a}, D_{b}, D_{c},\ldots$; Then we sample subsets from each subdomain using sample sizes $n_{1}, n_{2}, n_{3},\dots$ in ascending order. The resulting subsets are denoted  $D_{a}^{n_{1}}, D_{a}^{n_{2}}, \dots, D_{b}^{n_{1}}, D_{b}^{n_{2}},\ldots$. We train QA systems on each subset $D_{a}^{n_{1}}, D_{a}^{n_{2}}, \ldots$. The QA system trained on $D_{a}^{n_{1}}$ is denoted  $QA_{a}^{n_{1}}$. We evaluate each QA system on the test data $T$ which is also split into disjoint subdomains $T_{a}, T_{b}, T_{c},\ldots$ similar to the training data $D$.

\subsection{Question Type}

\begin{table}





\footnotesize
\setlength{\tabcolsep}{2pt}
  \centering
  \scalebox{1.05}{
  \begin{tabular}{p{1.5cm}cccccc}
    \toprule
    
     & & LOC & ENTY  &  HUM &  NUM &  DESC  \\
     
    \midrule
    
    \multirow{2}{4em}{\centering SQuAD1.1} & Train set & 11.4 & 27.6 & 20.7 & 24.5 & 15.5 \\
 & Dev set & 10.5 & 27.6 & 21.0 & 23.0 & 17.4 \\ \hline
 
     \multirow{2}{4em}{\centering NewsQA} & Train set & 11.4 & 16.9 & 30.0 & 18.8 & 22.6 \\
 & Dev set & 12.3 & 16.9 & 32.2 & 17.8 & 20.5 \\
    
    
    \bottomrule
    \end{tabular}%
    }
    \caption{The percentage(\%) of question types in the SQuAD1.1 and NewsQA train and dev sets.
}
  \label{tbl_0_percentage_qc}%

\end{table}

In this experiment, we investigate how QA models learn from QA examples with different question types. We adopt question classification data~\citep{li-roth-2002-learning_question_classification} to train a question classifier that categorizes questions into the following five classes: \textit{HUM, LOC, ENTY, DESC, NUM}~\footnote{Definitions and examples provided in the Appendix.}\citep{zhang2003question}. The QA examples in the training data are then partitioned into five categories according to their question type. Question type proportions for SQuAD1.1 and NewsQA are shown in Table~\ref{tbl_0_percentage_qc}, with a high proportion of \textit{ENTY} and \textit{NUM} questions in SQuAD1.1, while NewsQA has more \textit{HUM} and \textit{DESC} questions. We use QA examples of each question type to train a QA system increasing the training set size in intervals of 500 from 500 to 8000. We evaluate it on the test data, which is also divided into five categories according to question type.

\begin{figure}
    \centering
    \includegraphics[width=\linewidth]{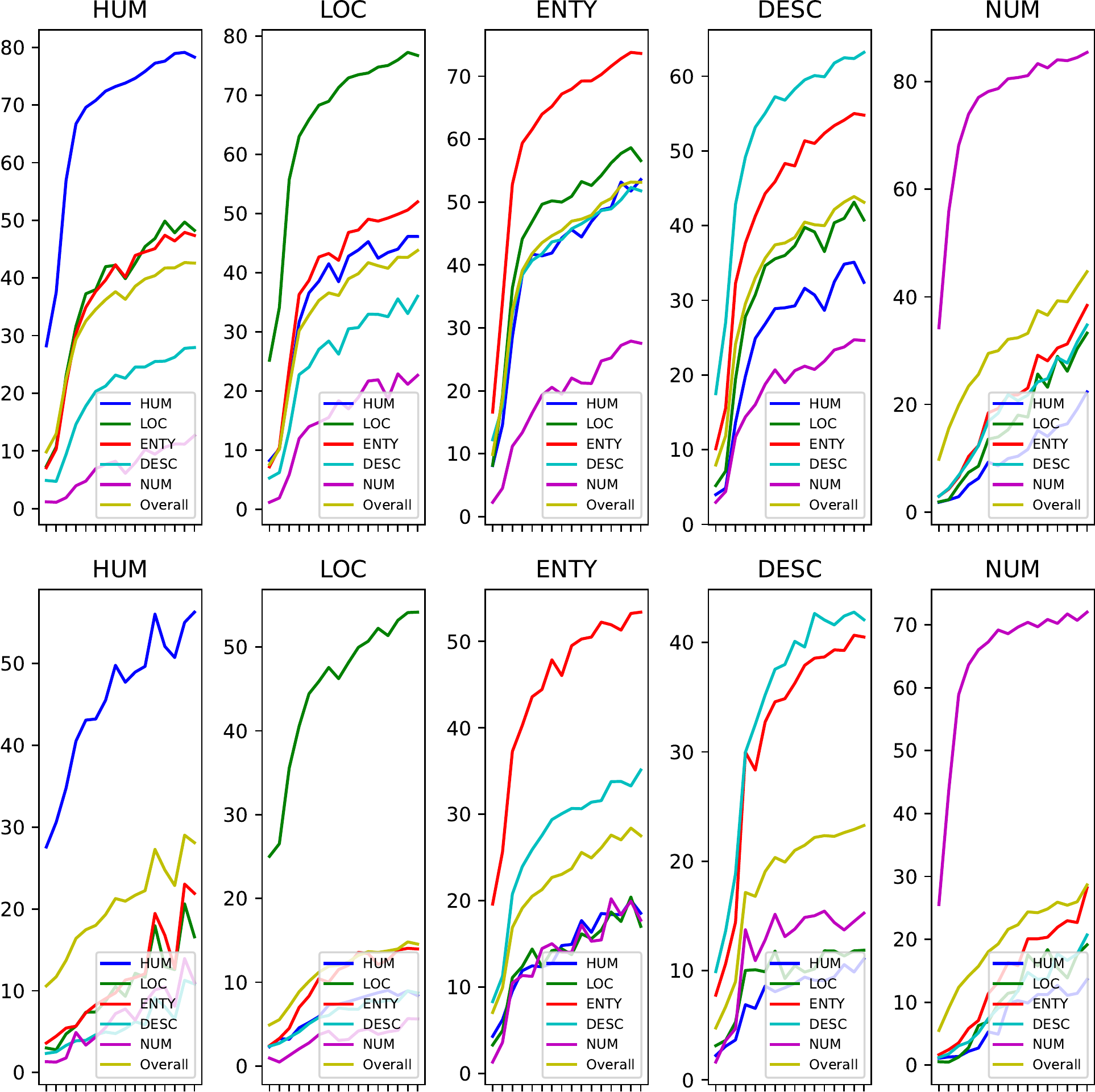}
    \caption{Visualization of F-1 learning curves for the QA systems trained on the \textit{subdomains} of five question types (\textit{HUM,LOC,ENTY,DESC,NUM}), tested on the \textit{subdomains} for each question type and the original dev set of SQuAD1.1 (top) and NewsQA (bottom).}
    \label{fig:qtype_f1_merge_dataset}
\end{figure}


\begin{figure*}
    \centering
    \includegraphics[scale=0.5,width=14cm]{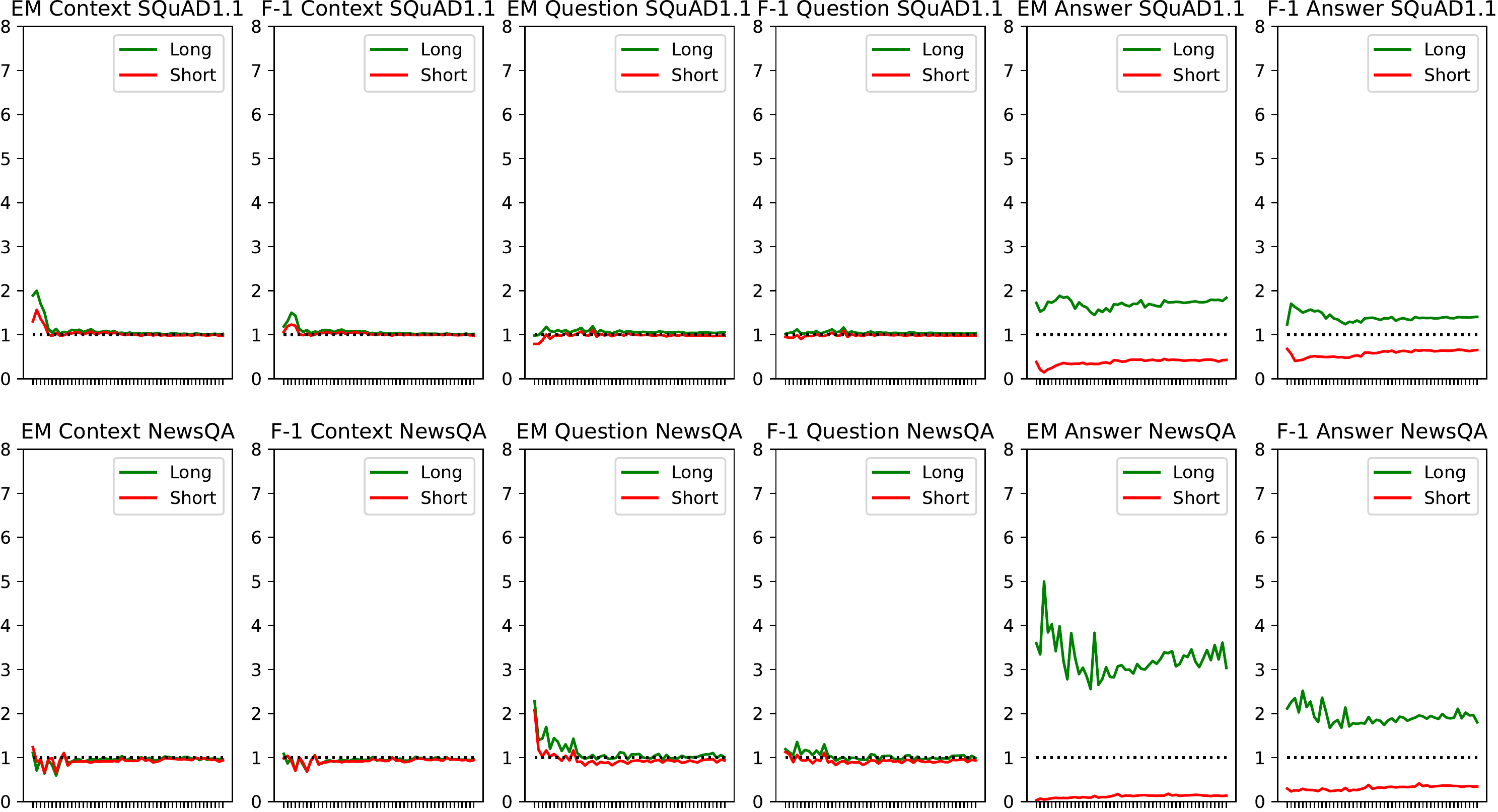}
    \caption{Visualization of performance~(EM and F-1 score) ratio curves over \textit{long} and \textit{short} context, question and answer~(from left to right) on SQuAD1.1 (top) and NewsQA (bottom). The \textit{green, red} lines represent the ratio of the performance on \textit{long group} and \textit{short group}. The dashed line is 1, indicating that two QA systems have the same performance. When the sample size increases, curves in \textit{context} and \textit{question} length converge to the dashed line, whereas there are substantial differences in the performance of $QA_{L}$ and $QA_{S}$ in \textit{answer length} subdomain.}
    \label{fig:text_length}
\end{figure*}

A visualisation of the resulting F-1 scores of the QA systems trained on each question type \textit{subdomain} is shown in  Figure~\ref{fig:qtype_f1_merge_dataset}, for both SQuAD1.1 and NewsQA.
The x-axis represents the training set size, the y-axis is the F-1 score. The results show that a QA system learns to answer a certain type of question mainly from the  examples of the same question type -- this is particularly true for \textit{HUM} and \textit{NUM} questions in SQuAD1.1 and \textit{HUM}, \textit{LOC} and \textit{NUM} questions in NewsQA. Taking \textit{NUM} questions as an example,  the rightmost plots in Figure~\ref{fig:qtype_f1_merge_dataset} show that  performance on other question types 
only results in a minor improvement as the training set size increases compared to performance improvements on that question type (\textit{NUM}). In other words, the QA system gets most of the knowledge it needs to answer \textit{NUM} questions  from the \textit{NUM} training examples and a similar pattern is also present for other question types. The results in Figure~\ref{fig:qtype_f1_merge_dataset} show that the subdomains defined by \textit{question type} is a source of bias when training and employing QA systems.
We suspect that word use and narrative style varies over question types, which inject bias into QA systems when learning from QA examples with different question types. Therefore, we need to improve the diversity of question types when constructing and organising QA data.

\subsection{Text Length}

\begin{figure*}
    \centering
    \includegraphics[scale=0.45,width=14cm]{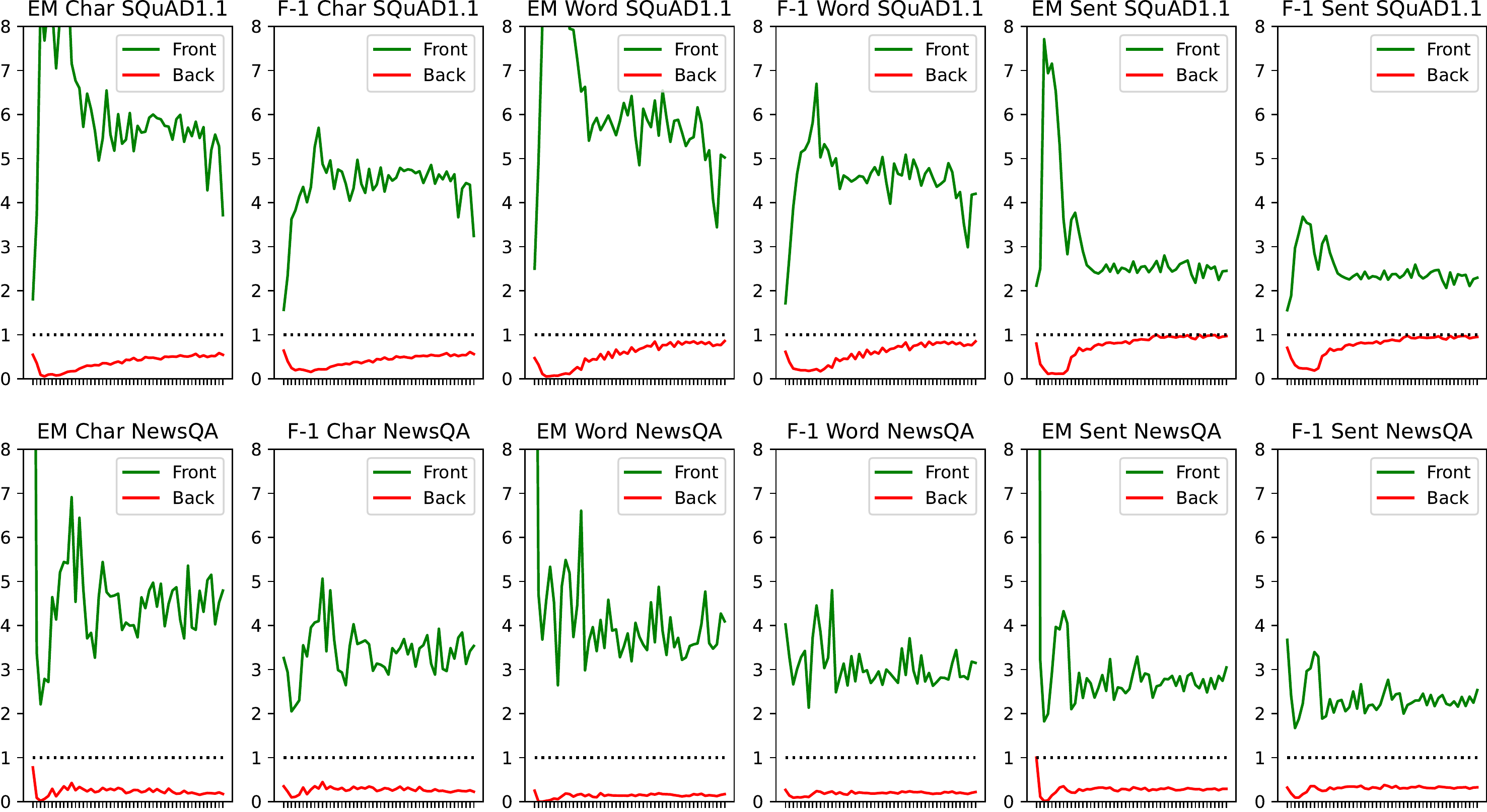}
    \caption{Visualization of performance~(EM and F-1 score) ratio curves over \textit{front} and \textit{back} answer positions (char-level, word-level and sentence-level from left to right) on SQuAD1.1 (top) and NewsQA (bottom). The \textit{green, red} lines represent the ratio of the performance on \textit{front group} and \textit{back group}. The dashed line is 1, indicating that two QA systems have the same performance. The curves show that there are substantially difference in the performance of $QA_{F}$ and $QA_{B}$ in \textit{answer position} subdomains especially for character-level and word-level answer positions.}
    \label{fig:answer_position}
\end{figure*}

The effect of text length to the performance and generalizability of neural models has been discussed in text classification and machine translation~\cite{amplayo2019text, varis-bojar-2021-sequence}. As for QA,  there are three components in a QA example: \textit{context, question, answer}. The length of each component could potentially introduce additional bias and affect how QA systems learn from QA data. For example, a short context could be \textit{easy} since a shorter context could reduce the search space for QA models to locate the answer; on the other hand, a short context could be \textit{hard} as shorter context contains less information to find the answer.  Therefore, the following question arises naturally: are \textit{short} and \textit{long} context/question/answer are equivalent? 

To answer this question, we split the QA datasets into \textit{short} and \textit{long} groups according to the median of the length of \textit{context, question, answer}~\footnote{See more statistics in the Appendix.}. Then we train QA systems on the QA examples sampled from \textit{short} ($QA_{S, context}, QA_{S, question}, QA_{S, answer}$) and \textit{long} ($QA_{L, context}, QA_{L, question}, QA_{L, answer}$) groups respectively, again increasing the training set size in intervals of  500 from 500 to 25000. The results are shown in Figure~\ref{fig:text_length}, where the x-axis is the data size and the y-axis is the ratio of the performance~(EM and F-1 score) of the $QA_{S}$ and corresponding $QA_{L}$ systems on the \textit{text length} subdomains of \textit{context, question, answer}. 
If $QA_{L}$ and $QA_{S}$ have no obvious difference in terms of performance on \textit{long} and \textit{short} groups respectively, the ratio of their performance should be close to $1$. The results show that the performances of $QA_{L}$ and $QA_{S}$ trained on the subdomains of \textit{context} and \textit{question} length have no obvious difference as all the three curves converge to $1$, although there are fluctuations when the sample sizes are small. In contrast,  $QA_{L}$ and $QA_{S}$ trained on the subdomains of \textit{answer} length behave differently -- see the subplots in the two rightmost columns of Figure~\ref{fig:text_length}.  $QA_{L}$ performs much better than $QA_{S}$ on the test examples with \textit{long}  answers and $QA_{L}$ performs much worse than $QA_{S}$ on the test examples with \textit{short} answers.
The results in Figure~\ref{fig:text_length} show that the length of answer introduces strong bias to QA systems. 
We think this stems from the fact that $QA_{L}$ tends to predict longer answers, whereas $QA_{S}$ tends to predict shorter answers, and they thus underperform in the counterpart \textit{subdomain}. We show the average length of the predicted answers of QA systems trained on \textit{long} and \textit{short} groups in Table~\ref{tbl_1_avg_langth_of_predicted_answer}.  Therefore, it is important to control the length distribution of answers when constructing and organising QA dataset especially using NER tools in the answer extraction phase since they tend to find shorter answers.

\begin{table}
\footnotesize
\setlength{\tabcolsep}{2pt}
  \centering
  \scalebox{1.05}{
  \begin{tabular}{lcccccccccc}
    \toprule
    &&  \multicolumn{2}{c}{Context} && \multicolumn{2}{c}{Question}  && \multicolumn{2}{c}{Answer} \\ [1ex]
    
     && Long &  Short  &&  Long &  Short &&  Long &  Short  \\
     
    \midrule
    
    SQuAD1.1 && 4.03 & 4.13 && 4.00  & 4.23 && 6.41  & 2.78 \\

    NewsQA && 5.46 & 5.33 && 5.16  & 5.87 && 9.57  & 3.51  \\
    
    \bottomrule
    \end{tabular}%
    }
    \caption{The average length of predicted answers of QA systems trained on \textit{long} and \textit{short} subdomains of \textit{context, question and answer} on SQuAD1.1 and NewsQA.
    }
  \label{tbl_1_avg_langth_of_predicted_answer}%
\end{table}

\subsection{Answer Position}
\newcite{ko-etal-2020-look-first-sent} study the effect of sentence-level answer position. Building on their analysis, 
we study the effect of two more types of answer position: 
character-level position and  word-level position.
We split the training set into \textit{front} and \textit{back} groups based on the median of the answer start positions at the  character, word and sentence level~\footnote{See more statistics in the Appendix.}. Then we train QA systems on the examples sampled from \textit{front} ($QA_{F,char}, QA_{F,word}, QA_{F,sent}$) and \textit{back} ($QA_{B,char}, QA_{B,word}, QA_{B,sent}$) groups respectively , increasing the training set size in intervals of 500 from 500 to 25000. We show the  results in Figure~\ref{fig:answer_position}, where the x-axis is the training set size and  the y-axis is the ratio of the performance~(EM and F-1 score) of $QA_{F}$ and $QA_{B}$ on the  \textit{answer position} subdomains at the character, word and sentence level.  

The results show that \textit{answer position} on all three levels are a source of bias. 
$QA_{F}$ performs much better than $QA_{B}$ on test instances with answer positions in the \textit{front}, whereas $QA_{B}$ performs much better than $QA_{F}$ on test instances with answer positions at the \textit{back}. The effect of bias is more serious at the character and word level.
We think this answer position bias is happening because words in different positions have different position embeddings, which could also affect word semantics in transformer architectures~\cite{transformer, wang2020position}. This suggests the need to make sure answer position distribution is balanced as well as the need to develop QA systems that are more robust to answer position variation.

\section{Discussion and Conclusion}

We presented a series of
experiments investigating the \textit{out-of-subdomain} performance of QA systems on two popular English extractive QA datasets: SQuAD1.1 and NewsQA. The experimental results show that the \textit{subdomains} defined by \textit{question type, answer length, answer position} inject strong bias into QA systems, with the result that the performance of QA systems is negatively impacted when train and test data come from different \textit{subdomains}. The experiments provide useful information on how to control question diversity, answer length distribution as well as the answer positions when constructing QA datasets and employing QA systems. 
In future work, we aim to apply our analysis to multilingual data to explore how QA models behave across different languages and we plan to investigate
other types of QA beyond extractive QA.

\section*{Acknowledgements}
This work was funded by Science Foundation Ireland through the SFI Centre for Research Training in Machine Learning (18/CRT/6183). We thank James Barry for his proofreading and helpful comments. We also thank the reviewers for their helpful comments.
\bibliography{anthology,custom}
\bibliographystyle{acl_natbib}

\appendix

\section{Appendix}
\label{sec:appendix}
\subsection{Experimental Setup}
We use bert-based-uncased as our QA model, the learning rate is set to 3e-5, the maximum sequence length is set to 384, the doc stride length is set to 128, we run the training process for 2 epochs for training each QA system in experiments, the training was conducted on one GeForce GTX 3090 GPU, the training batch size is 48.

\subsection{Average Text Length and Answer Position for All Question Types}

We show the average text length of \textit{context, question and answer} as well as the average answer position on character-level, word-level and sentence-level for QA examples in all question types in SQuAD1.1 and NewsQA in Table~\ref{tbl_3_text_len_qtype} and Table~\ref{tbl_3_ans_pos_qtype}. 

\begin{table}
\footnotesize
\setlength{\tabcolsep}{2pt}
  \centering
  \scalebox{1.05}{
  \begin{tabular}{p{1.5cm}cccc}
    \toprule
    
     & & Context & Question  &  Answer \\
     
    \midrule
    
    \multirow{5}{4em}{\centering SQuAD1.1} 
  & HUM & 123.20 & 9.79 & 2.82 \\
  & LOC & 117.18 & 9.62 & 2.78 \\ 
  & DESC & 119.32 & 9.91 & 5.82\\
  & ENTY & 117.43 & 10.54 & 3.04 \\
  & NUM & 121.09 & 10.11 & 2.08 \\
 
 \hline
 
     \multirow{5}{4em}{\centering NewsQA} 
    & HUM & 495.79 & 6.55 & 2.82 \\
    & LOC & 478.84 & 6.34 & 2.87 \\ 
    & DESC & 513.00 & 6.25 & 7.62 \\
    & ENTY & 505.94 & 6.76 & 4.27 \\
    & NUM & 476.23 & 7.20 & 2.07 \\
    
    \bottomrule
    \end{tabular}%
    }
    \caption{The average text length of context, question and answer in QA examples of each question type in the SQuAD1.1 and NewsQA training data.}
  \label{tbl_3_text_len_qtype}%

\end{table}

\begin{table}
\footnotesize
\setlength{\tabcolsep}{2pt}
  \centering
  \scalebox{1.05}{
  \begin{tabular}{p{1.5cm}cccc}
    \toprule
    
     & & Char-Level & Word-Level  &  Sent-Level \\
     
    \midrule
    
    \multirow{5}{4em}{\centering SQuAD1.1} 
  & HUM & 317.85 & 54.90 & 1.61 \\
  & LOC & 308.81 & 53.71 & 1.53 \\ 
  & DESC & 342.97 & 60.00 & 1.79\\
  & ENTY & 317.75 & 55.16 & 1.63 \\
  & NUM & 315.78 & 56.19 & 1.67 \\
 
 \hline
 
     \multirow{5}{4em}{\centering NewsQA} 
    & HUM & 532.11 & 101.02 & 3.71 \\
    & LOC & 566.02 & 107.99 & 3.95 \\ 
    & DESC & 844.13 & 160.05 & 5.98 \\
    & ENTY & 751.48 & 143.90 & 5.49 \\
    & NUM & 763.73 & 145.26 & 5.47 \\
    
    \bottomrule
    \end{tabular}%
    }
    \caption{The average answer position of character-level, word-level and sentence-level in QA examples of each question type in the SQuAD1.1 and NewsQA training data.}
  \label{tbl_3_ans_pos_qtype}%

\end{table}

\begin{table}
\footnotesize
\setlength{\tabcolsep}{2pt}
  \centering
  \scalebox{1.05}{
  \begin{tabular}{p{1.5cm}ccc}
    \toprule
    
     & Context & Question  & Answer \\ 
 \midrule 

SQuAD1.1 & 110 & 10 & 2 \\
 \hline 
 NewsQA & 534 & 6 & 2 \\
    
    \bottomrule
    \end{tabular}%
    }
    \caption{The median of the \textit{context, question, answer} length used to partition \textit{long} and \textit{short} subdomains.}
  \label{tbl_0_median_text_len}%

\end{table}

\begin{table}
\footnotesize
\setlength{\tabcolsep}{2pt}
  \centering
  \scalebox{1.05}{
  \begin{tabular}{p{1.5cm}cccccc}
    \toprule
    
     & & LOC & ENTY  & HUM & NUM & DESC \\ 
 \midrule 

 \multirow{2}{4em}{\centering SQuAD1.1} 
& Long & 11.11 & 26.68 & 21.65 & 24.8 & 15.43 \\
& Short & 11.73 & 28.42 & 19.68 & 24.2 & 15.52 \\
 \hline 
 \multirow{2}{4em}{\centering NewsQA} 
& Long & 10.4 & 18.08 & 29.94 & 16.81 & 24.71 \\
& Short & 12.38 & 15.86 & 30.24 & 20.9 & 20.55 \\
    
    \bottomrule
    \end{tabular}%
    }
    \caption{The percentage of each question type in \textit{long context} and \textit{short context} groups.}
  \label{tbl_1_1_text_len_qtype}%

\end{table}

\begin{table}
\footnotesize
\setlength{\tabcolsep}{2pt}
  \centering
  \scalebox{1.05}{
  \begin{tabular}{p{1.5cm}cccccc}
    \toprule
    
     & & LOC & ENTY  & HUM & NUM & DESC \\
 \midrule

 \multirow{2}{4em}{\centering SQuAD1.1}
& Long & 10.36 & 28.59 & 20.37 & 24.73 & 15.63 \\
& Short & 12.11 & 26.90 & 20.84 & 24.35 & 15.37 \\
 \hline
 \multirow{2}{4em}{\centering NewsQA}
& Long & 9.45 & 18.29 & 29.70 & 23.66 & 18.90 \\
& Short & 12.96 & 15.91 & 30.40 & 14.98 & 25.63 \\
    
    \bottomrule
    \end{tabular}%
    }
    \caption{The percentage of each question type in \textit{long question} and \textit{short question} groups.}
  \label{tbl_1_2_text_len_qtype}%

\end{table}

\begin{table}
\footnotesize
\setlength{\tabcolsep}{2pt}
  \centering
  \scalebox{1.05}{
  \begin{tabular}{p{1.5cm}cccccc}
    \toprule
    
     & & LOC & ENTY  & HUM & NUM & DESC \\
 \midrule

 \multirow{2}{4em}{\centering SQuAD1.1}
& Long & 10.87 & 27.32 & 19.69 & 21.8 & 19.86 \\
& Short & 11.79 & 27.72 & 21.29 & 26.29 & 12.55 \\
 \hline
 \multirow{2}{4em}{\centering NewsQA}
& Long & 9.37 & 19.87 & 23.16 & 9.31 & 38.17 \\
& Short & 13.13 & 14.48 & 36.03 & 27.05 & 9.29 \\
    
    \bottomrule
    \end{tabular}%
    }
    \caption{The percentage of each question type in \textit{long answer} and \textit{short answer} groups.}
  \label{tbl_1_3_text_len_qtype}%

\end{table}

\begin{table}
\footnotesize
\setlength{\tabcolsep}{2pt}
  \centering
  \scalebox{1.05}{
  \begin{tabular}{p{1.5cm}cccc}
    \toprule
    
     & & Context & Question  & Answer \\ 
 \midrule 

 \multirow{2}{4em}{\centering SQuAD1.1} 
& Long & 84.53 & 9.99 & 3.09 \\
& Short & 155.88 & 10.14 & 3.23 \\
 \hline 
 \multirow{2}{4em}{\centering NewsQA} 
& Long & 350.44 & 6.54 & 3.79 \\
& Short & 641.35 & 6.69 & 4.25 \\
    
    \bottomrule
    \end{tabular}%
    }
    \caption{The average answer position on character-level, word-level and sentence-level in QA examples of \textit{long context} and \textit{short context} groups.}
  \label{tbl_1_1_text_len_by_text_len}%

\end{table}

\begin{table}
\footnotesize
\setlength{\tabcolsep}{2pt}
  \centering
  \scalebox{1.05}{
  \begin{tabular}{p{1.5cm}cccc}
    \toprule
    
     & & Context & Question  & Answer \\ 
 \midrule 

 \multirow{2}{4em}{\centering SQuAD1.1} 
& Long & 119.12 & 7.8 & 3.25 \\
& Short & 120.76 & 13.57 & 3.03 \\
 \hline 
 \multirow{2}{4em}{\centering NewsQA} 
& Long & 491.15 & 4.96 & 4.45 \\
& Short & 501.55 & 8.66 & 3.49 \\
    
    \bottomrule
    \end{tabular}%
    }
    \caption{The average answer position on character-level, word-level and sentence-level in QA examples of \textit{long question} and \textit{short question} groups.}
  \label{tbl_1_2_text_len_by_text_len}%

\end{table}

\begin{table}
\footnotesize
\setlength{\tabcolsep}{2pt}
  \centering
  \scalebox{1.05}{
  \begin{tabular}{p{1.5cm}cccc}
    \toprule
    
     & & Context & Question  & Answer \\ 
 \midrule 

 \multirow{2}{4em}{\centering SQuAD1.1} 
& Long & 119.08 & 10.18 & 1.42 \\
& Short & 120.79 & 9.88 & 5.77 \\
 \hline 
 \multirow{2}{4em}{\centering NewsQA} 
& Long & 489.32 & 6.82 & 1.5 \\
& Short & 503.34 & 6.37 & 6.95 \\
    
    \bottomrule
    \end{tabular}%
    }
    \caption{The average answer position on character-level, word-level and sentence-level in QA examples of \textit{long answer} and \textit{short answer} groups.}
  \label{tbl_1_3_text_len_by_text_len}%

\end{table}

\begin{table}
\footnotesize
\setlength{\tabcolsep}{2pt}
  \centering
  \scalebox{1.05}{
  \begin{tabular}{p{1.5cm}cccc}
    \toprule
    
     & & Char & Word  & Sent \\ 
 \midrule 

 \multirow{2}{4em}{\centering SQuAD1.1} 
& Long & 402.02 & 70.36 & 2.14 \\
& Short & 239.75 & 41.78 & 1.17 \\
 \hline 
 \multirow{2}{4em}{\centering NewsQA} 
& Long & 864.85 & 165.73 & 6.40 \\
& Short & 510.58 & 95.94 & 3.37 \\
    
    \bottomrule
    \end{tabular}%
    }
    \caption{The average answer position on character-level, word-level and sentence-level in QA examples of \textit{long context} and \textit{short context} groups.}
  \label{tbl_1_1_ans_pos_by_text_len}%

\end{table}

\begin{table}
\footnotesize
\setlength{\tabcolsep}{2pt}
  \centering
  \scalebox{1.05}{
  \begin{tabular}{p{1.5cm}cccc}
    \toprule
    
     & & Char & Word  & Sent \\
 \midrule

 \multirow{2}{4em}{\centering SQuAD1.1}
& Long & 342.02 & 59.70 & 1.74 \\
& Short & 305.65 & 53.45 & 1.58 \\
 \hline
 \multirow{2}{4em}{\centering NewsQA}
& Long & 726.78 & 138.64 & 5.22 \\
& Short & 655.98 & 124.50 & 4.61 \\
    
    \bottomrule
    \end{tabular}%
    }
    \caption{The average answer position on character-level, word-level and sentence-level in QA examples of \textit{long question} and \textit{short question} groups.}
  \label{tbl_1_2_ans_pos_by_text_len}%

\end{table}

\begin{table}
\footnotesize
\setlength{\tabcolsep}{2pt}
  \centering
  \scalebox{1.05}{
  \begin{tabular}{p{1.5cm}cccc}
    \toprule
    
      & & Char & Word  & Sent \\
 \midrule

 \multirow{2}{4em}{\centering SQuAD1.1}
& Long & 324.65 & 57.77 & 1.71 \\
& Short & 316.70 & 54.65 & 1.60 \\
 \hline
 \multirow{2}{4em}{\centering NewsQA}
& Long & 795.46 & 150.20 & 5.61 \\
& Short & 595.00 & 114.17 & 4.26 \\
    
    \bottomrule
    \end{tabular}%
    }
    \caption{The average answer position on character-level, word-level and sentence-level in QA examples of \textit{long answer} and \textit{short answer} groups.}
  \label{tbl_1_3_ans_pos_by_text_len}%

\end{table}
\subsection{Question Type Proportions, Average Text Length and Average Answer Position for \textit{Long} and \textit{Short} Text Length}

The median of the \textit{context, question, answer} is shown in Table~\ref{tbl_0_median_text_len}. We show the question type proportion, average text length for \textit{context, question and answer}  as well as the average answer position on character-level, word-level and sentence-level for QA examples in \textit{long} and \textit{short} groups of \textit{context, question, answer} in SQuAD1.1 and NewsQA in Table~\ref{tbl_1_1_text_len_qtype}, Table~\ref{tbl_1_2_text_len_qtype}, Table~\ref{tbl_1_3_text_len_qtype}, 
Table~\ref{tbl_1_1_text_len_by_text_len}
Table~\ref{tbl_1_2_text_len_by_text_len}, Table~\ref{tbl_1_3_text_len_by_text_len}, 
Table~\ref{tbl_1_1_ans_pos_by_text_len}, 
Table~\ref{tbl_1_2_ans_pos_by_text_len}, 
Table~\ref{tbl_1_3_ans_pos_by_text_len}.

\subsection{Question Type Proportions, Average Text Length and Average Answer Position for QA examples with \textit{Front} and \textit{Back} Answer Positions}

The median of the answer position on character-level, word-level and sentence-level is shown in Table~\ref{tbl_0_median_ans_pos}. We show the question type proportion, average text length for \textit{context, question and answer} as well as the average answer position on character-level, word-level and sentence-level for QA examples in \textit{front} and \textit{back} groups of answer positions in character-level, word-level and sentence-level in SQuAD1.1 and NewsQA in Table~\ref{tbl_1_1_ans_pos_qtype}, Table~\ref{tbl_1_2_ans_pos_qtype}, Table~\ref{tbl_1_3_ans_pos_qtype},
Table~\ref{tbl_1_1_ans_pos_by_ans_pos}, Table~\ref{tbl_1_2_ans_pos_by_ans_pos}, Table~\ref{tbl_1_3_ans_pos_by_ans_pos}, 
Table~\ref{tbl_1_1_text_len_by_ans_pos}, 
Table~\ref{tbl_1_2_text_len_by_ans_pos}, 
Table~\ref{tbl_1_3_text_len_by_ans_pos}.

\begin{table}
\footnotesize
\setlength{\tabcolsep}{2pt}
  \centering
  \scalebox{1.05}{
  \begin{tabular}{p{1.5cm}ccc}
    \toprule
    
     & Char & Word  & Sent \\ 
 \midrule 

SQuAD1.1 & 262 & 46 & 1 \\
 \hline 
 NewsQA & 358 & 67 &  2\\
    
    \bottomrule
    \end{tabular}%
    }
    \caption{The median of the answer position on character-level, word-level and sentence-level used to partition \textit{front} and \textit{back} subdomains.}
  \label{tbl_0_median_ans_pos}%

\end{table}

\begin{table}
\footnotesize
\setlength{\tabcolsep}{2pt}
  \centering
  \scalebox{1.05}{
  \begin{tabular}{p{1.5cm}cccccc}
    \toprule
    
     & & LOC & ENTY  & HUM & NUM & DESC \\
 \midrule

 \multirow{2}{4em}{\centering SQuAD1.1}
& Front & 11.74 & 27.8 & 20.25 & 24.97 & 14.81 \\
& Back & 11.11 & 27.32 & 21.06 & 24.02 & 16.14 \\
 \hline
 \multirow{2}{4em}{\centering NewsQA}
& Front & 13.07 & 15.59 & 37.2 & 15.61 & 18.46 \\
& Back & 9.71 & 18.36 & 22.97 & 22.1 & 26.8 \\
    
    \bottomrule
    \end{tabular}%
    }
    \caption{The percentage of each question type in \textit{front} and \textit{back} groups on character-level answer position}
  \label{tbl_1_1_ans_pos_qtype}%

\end{table}

\begin{table}
\footnotesize
\setlength{\tabcolsep}{2pt}
  \centering
  \scalebox{1.05}{
  \begin{tabular}{p{1.5cm}cccccc}
    \toprule
    
     & & LOC & ENTY  & HUM & NUM & DESC \\
 \midrule

 \multirow{2}{4em}{\centering SQuAD1.1}
& Front & 11.76 & 28.05 & 20.28 & 24.49 & 14.99 \\
& Back & 11.16 & 27.08 & 21.00 & 24.45 & 15.94 \\
 \hline
 \multirow{2}{4em}{\centering NewsQA}
& Front & 13.02 & 15.59 & 37.20 & 15.64 & 18.48 \\
& Back & 9.74 & 18.43 & 22.85 & 22.11 & 26.81 \\
    
    \bottomrule
    \end{tabular}%
    }
    \caption{The percentage of each question type in \textit{front} and \textit{back} groups on word-level answer position}
  \label{tbl_1_2_ans_pos_qtype}%

\end{table}

\begin{table}
\footnotesize
\setlength{\tabcolsep}{2pt}
  \centering
  \scalebox{1.05}{
  \begin{tabular}{p{1.5cm}cccccc}
    \toprule
    
     & & LOC & ENTY  & HUM & NUM & DESC \\
 \midrule

 \multirow{2}{4em}{\centering SQuAD1.1}
& Front & 11.72 & 27.83 & 20.60 & 24.48 & 14.95 \\
& Back & 11.04 & 27.18 & 20.71 & 24.56 & 16.15 \\
 \hline
 \multirow{2}{4em}{\centering NewsQA}
& Front & 13.19 & 15.76 & 36.08 & 16.36 & 18.54 \\
& Back & 9.56 & 18.54 & 23.11 & 22.06 & 26.67 \\
    
    \bottomrule
    \end{tabular}%
    }
    \caption{The percentage of each question type in \textit{front} and \textit{back} groups on sentence-level answer position}
  \label{tbl_1_3_ans_pos_qtype}%

\end{table}

\begin{table}
\footnotesize
\setlength{\tabcolsep}{2pt}
  \centering
  \scalebox{1.05}{
  \begin{tabular}{p{1.5cm}cccc}
    \toprule
    
     & & Char & Word  & Sent \\ 
 \midrule 

 \multirow{2}{4em}{\centering SQuAD1.1} 
& Front & 116.25 & 20.6 & 0.44 \\
& Back & 524.15 & 91.3 & 2.85 \\
 \hline 
 \multirow{2}{4em}{\centering NewsQA} 
& Front & 145.24 & 28.72 & 0.61 \\
& Back & 1230.24 & 232.96 & 9.15 \\
    
    \bottomrule
    \end{tabular}%
    }
    \caption{The average answer position on character-level, word-level and sentence-level in QA examples of \textit{front} and \textit{back} groups of character-level answer position.}
  \label{tbl_1_1_ans_pos_by_ans_pos}%

\end{table}

\begin{table}
\footnotesize
\setlength{\tabcolsep}{2pt}
  \centering
  \scalebox{1.05}{
  \begin{tabular}{p{1.5cm}cccc}
    \toprule
    
      & & Char & Word  & Sent \\ 
 \midrule 

 \multirow{2}{4em}{\centering SQuAD1.1} 
& Front & 127.4 & 19.34 & 0.44 \\
& Back & 515.71 & 93.09 & 2.88 \\
 \hline 
 \multirow{2}{4em}{\centering NewsQA} 
& Front & 151.46 & 28.04 & 0.65 \\
& Back & 1229.77 & 234.74 & 9.17 \\
    
    \bottomrule
    \end{tabular}%
    }
    \caption{The average answer position on character-level, word-level and sentence-level in QA examples of \textit{front} and \textit{back} groups of word-level answer position.}
  \label{tbl_1_2_ans_pos_by_ans_pos}%

\end{table}

\begin{table}
\footnotesize
\setlength{\tabcolsep}{2pt}
  \centering
  \scalebox{1.05}{
  \begin{tabular}{p{1.5cm}cccc}
    \toprule
    
     & & Char & Word  & Sent \\ 
 \midrule 

 \multirow{2}{4em}{\centering SQuAD1.1} 
& Front & 158.46 & 26.12 & 0.4 \\
& Back & 532.52 & 95.11 & 3.28 \\
 \hline 
 \multirow{2}{4em}{\centering NewsQA} 
& Front & 183.56 & 35.56 & 0.63 \\
& Back & 1280.56 & 242.86 & 9.89 \\
    
    \bottomrule
    \end{tabular}%
    }
    \caption{The average answer position on character-level, word-level and sentence-level in QA examples of \textit{front} and \textit{back} groups of sentence-level answer position.}
  \label{tbl_1_3_ans_pos_by_ans_pos}%

\end{table}

\begin{table}
\footnotesize
\setlength{\tabcolsep}{2pt}
  \centering
  \scalebox{1.05}{
  \begin{tabular}{p{1.5cm}cccc}
    \toprule
    
     & & Context & Question  & Answer \\
 \midrule

 \multirow{2}{4em}{\centering SQuAD1.1}
& Front & 108.80 & 9.83 & 3.06 \\
& Back & 130.77 & 10.30 & 3.26 \\
 \hline
 \multirow{2}{4em}{\centering NewsQA}
& Front & 473.52 & 6.50 & 3.28 \\
& Back & 518.08 & 6.72 & 4.75 \\
    
    \bottomrule
    \end{tabular}%
    }
    \caption{The average text length of context, question and answer in QA examples of \textit{front} and \textit{back} groups of character-level answer position}
  \label{tbl_1_1_text_len_by_ans_pos}%

\end{table}

\begin{table}
\footnotesize
\setlength{\tabcolsep}{2pt}
  \centering
  \scalebox{1.05}{
  \begin{tabular}{p{1.5cm}cccc}
    \toprule
    
     & & Context & Question  & Answer \\
\midrule

 \multirow{2}{4em}{\centering SQuAD1.1}
& Front & 109.21 & 9.84 & 3.03 \\
& Back & 130.50 & 10.28 & 3.30 \\
 \hline
 \multirow{2}{4em}{\centering NewsQA}
& Front & 473.13 & 6.50 & 3.32 \\
& Back & 518.72 & 6.72 & 4.72 \\
    
    \bottomrule
    \end{tabular}%
    }
    \caption{The average text length of context, question and answer in QA examples of \textit{front} and \textit{back} groups of word-level answer position}
  \label{tbl_1_2_text_len_by_ans_pos}%

\end{table}

\begin{table}
\footnotesize
\setlength{\tabcolsep}{2pt}
  \centering
  \scalebox{1.05}{
  \begin{tabular}{p{1.5cm}cccc}
    \toprule
    
     & & Context & Question  & Answer \\
 \midrule

 \multirow{2}{4em}{\centering SQuAD1.1}
& Front & 110.14 & 9.93 & 3.04 \\
& Back & 132.44 & 10.23 & 3.33 \\
 \hline
 \multirow{2}{4em}{\centering NewsQA}
& Front & 474.28 & 6.52 & 3.58 \\
& Back & 521.11 & 6.73 & 4.54 \\
    
    \bottomrule
    \end{tabular}%
    }
    \caption{The average text length of context, question and answer in QA examples of \textit{front} and \textit{back} groups of sentence-level answer position}
  \label{tbl_1_3_text_len_by_ans_pos}%

\end{table}

\begin{figure*}
    \centering
    \includegraphics[scale=0.45,width=14cm]{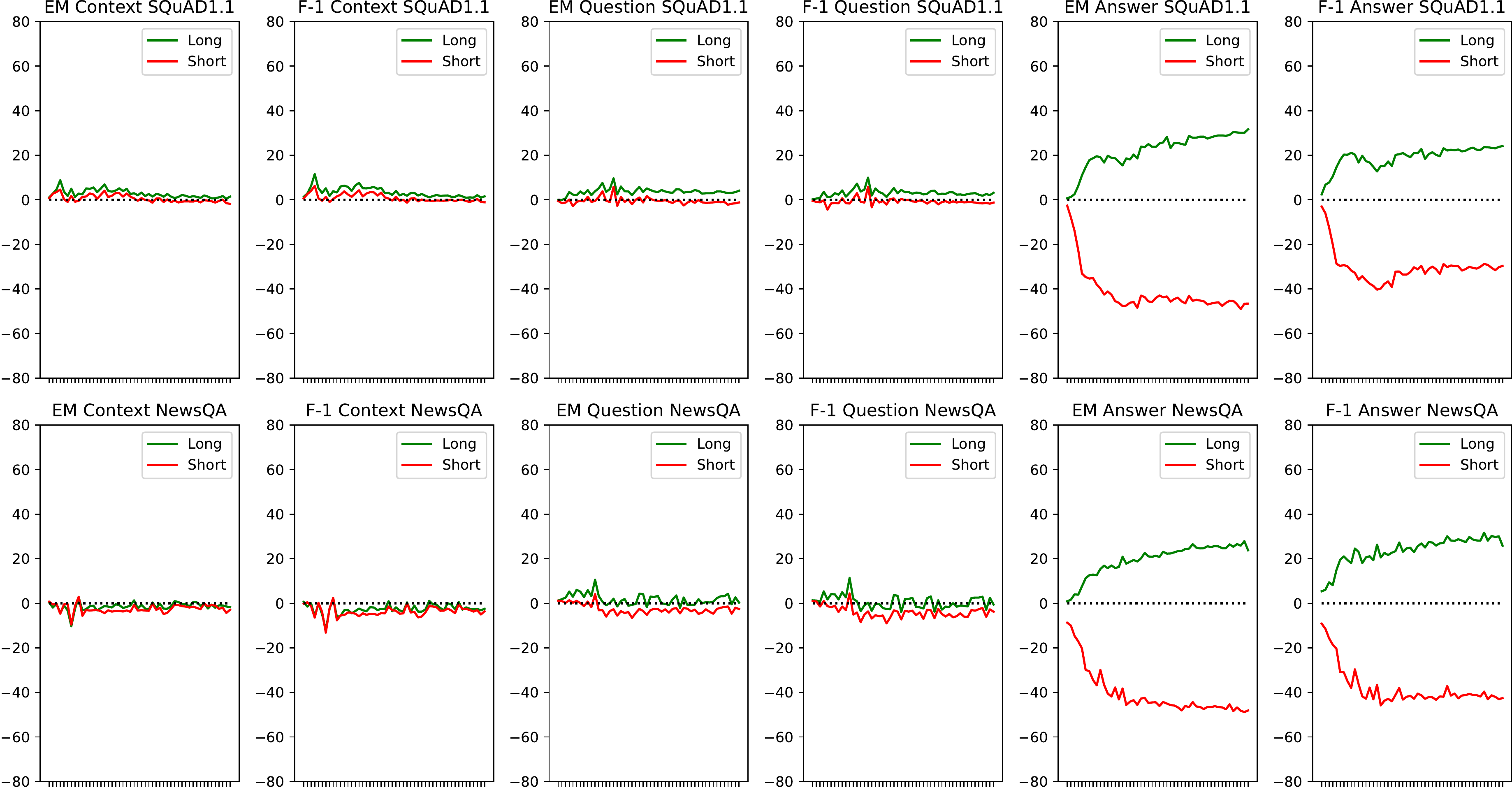}
    \caption{Visualization of performance~(EM and F-1 score) difference curves over \textit{short} and \textit{long} context, question and answer~(from left to right) on SQuAD1.1 (top) and NewsQA (bottom). The \textit{green, red} lines represent the difference of the performance on \textit{long group} and \textit{short group}. The dashed line is 0, indicating that two QA systems have the same performance. When the sample size increases, curves in \textit{context} and \textit{question} length converge to the dashed line, whereas there are substantial differences in the performance of $QA_{L}$ and $QA_{S}$ in \textit{answer length} subdomain.}
    \label{fig:text_len_appendix}
\end{figure*}

\begin{figure*}
    \centering
    \includegraphics[scale=0.45,width=14cm]{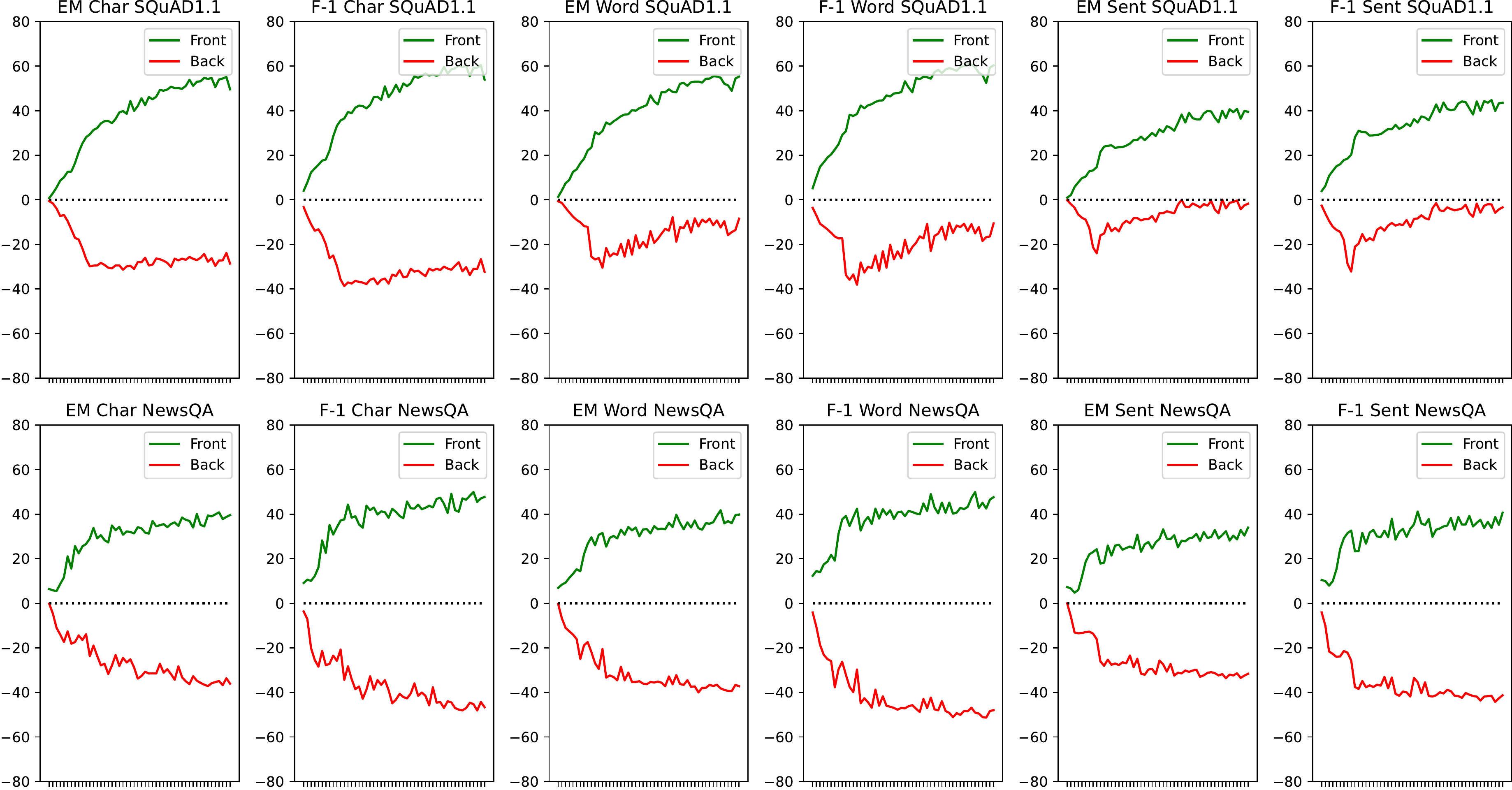}
    \caption{Visualization of performance~(EM and F-1 score) difference curves over \textit{front} and \textit{back} answer positions (char-level, word-level and sentence-level from left to right) on SQuAD1.1 (top) and NewsQA (bottom). The \textit{green, red} lines represent the difference of the performance on \textit{front group} and \textit{back group}. The dashed line is 0, indicating that two QA systems have the same performance. The curves show that there are substantially difference in the performance of $QA_{F}$ and $QA_{B}$ in \textit{answer position} subdomains especially for character-level and word-level answer positions.}
    \label{fig:answer_position_appendix}
\end{figure*}

\subsection{Question type definition and examples}

We show the definitions of question type \textit{HUM, LOC, ENTY, DESC, NUM} and some examples from the question classification data\footnote{https://cogcomp.seas.upenn.edu/Data/QA/QC/} and predictions of SQuAD1.1 and NewsQA in Table~\ref{question_type_examples}.
\begin{table*}
    \centering

  \begin{tabular}{|p{3cm}|p{4cm}|p{8cm}|}

    \hline
    
    Question type & Definition & Examples \\
    
    \hline
    
    \textit{HUM} & people, individual, group, title & 
    What contemptible scoundrel stole the cork from my lunch ? \newline
    Which professor sent the first wireless message in the USA ? \newline
    Who was sentenced to death in February ? \\

    \hline
    \textit{LOC} & location, city, country, mountain, state & 
    Where is the Kalahari desert ? \newline
    Where is the theology library at Notre Dame ? \newline
    Where was Cretan when he heard screams ? \\
    
    \hline
    \textit{ENTY} & animal, body, color, creation, currency, disease/medical, event, food, instrument, language, plant, product, religion, sport, symbol, technique, term, vehicle & 
    What relative of the racoon is sometimes known as the cat-bear ? \newline
    What is the world's oldest monographic music competition ? \newline
    What was the name of the film about Jack Kevorkian ? \\
    
    \hline
    \textit{DESC} & definition, description, manner, reason & 
    What is Eagle 's syndrome styloid process ? \newline
    How did Beyonce describe herself as a feminist ? \newline
    What are suspects blamed for ? \\
    
    \hline
    \textit{NUM} & code, count, date, distance, money, order, other, percent, period, speed, temperature, size, weight & 
    How many calories are there in a Big Mac ? \newline
    What year did Nintendo announce a new Legend of Zelda was in the works for Gamecube ? \newline
    How many tons of cereal did Kelloggs donate ? \\
    
    \hline
    
    \end{tabular}%
    \caption{Definition of each question type and corresponding examples in SQuAD1.1 and NewsQA.
      \label{question_type_examples}%
    }

\end{table*}

\subsection{QA examples with \textit{long} answers and \textit{short} answers}

We give some QA examples with \textit{long} answers and \textit{short} answers in Table~\ref{appendix:answer_len_examples_1} and Table~\ref{appendix:answer_len_examples_2}.

\begin{table*}
    \centering

  \begin{tabular}{|p{3cm}|p{4cm}|p{8cm}|}

    \hline
    
    Answer Length & Question & Context \\
    
    \hline
    
     Long &  Where was the main focus of Pan-Slavism? & Pan-Slavism, a movement which came into prominence in the mid-19th century, emphasized the common heritage and unity of all the Slavic peoples. The main focus was in the Balkans where the South Slavs had been ruled for centuries by other empires: \textbf{\textit{the Byzantine Empire, Austria-Hungary, the Ottoman Empire, and Venice}}. The Russian Empire used Pan-Slavism as a political tool; as did the Soviet Union, which gained political-military influence and control over most Slavic-majority nations between 1945 and 1948 and retained a hegemonic role until the period 1989–1991. \\ \hline

    Long & What is one reason for homologs to appear? & Genes with a most recent common ancestor, and thus a shared evolutionary ancestry, are known as homologs. These genes appear either from \textbf{\textit{gene duplication within an organism's genome}}, where they are known as paralogous genes, or are the result of divergence of the genes after a speciation event, where they are known as orthologous genes,:7.6 and often perform the same or similar functions in related organisms. It is often assumed that the functions of orthologous genes are more similar than those of paralogous genes, although the difference is minimal. \\

    \hline
    
    Long & How does the water vapor that rises in warm air turn into clouds? & Solar radiation is absorbed by the Earth's land surface, oceans – which cover about 71\% of the globe – and atmosphere. Warm air containing evaporated water from the oceans rises, causing atmospheric circulation or convection. \textbf{\textit{When the air reaches a high altitude, where the temperature is low, water vapor condenses into clouds}}, which rain onto the Earth's surface, completing the water cycle. The latent heat of water condensation amplifies convection, producing atmospheric phenomena such as wind, cyclones and anti-cyclones. Sunlight absorbed by the oceans and land masses keeps the surface at an average temperature of 14 °C. By photosynthesis green plants convert solar energy into chemically stored energy, which produces food, wood and the biomass from which fossil fuels are derived. \\ 
    \hline

    \end{tabular}%
    \caption{Examples of QA examples with \textit{long} answers where answers are highlighted.}
    \label{appendix:answer_len_examples_1}%

\end{table*}
\begin{table*}
    \centering

  \begin{tabular}{|p{3cm}|p{4cm}|p{8cm}|}

    \hline
    
    Answer Length & Question & Context \\
    
    \hline
    
     Short &  Who led the Exodus? & According to the Hebrew Bible narrative, Jewish ancestry is traced back to the Biblical patriarchs such as Abraham, Isaac and Jacob, and the Biblical matriarchs Sarah, Rebecca, Leah, and Rachel, who lived in Canaan around the 18th century BCE. Jacob and his family migrated to Ancient Egypt after being invited to live with Jacob's son Joseph by the Pharaoh himself. The patriarchs' descendants were later enslaved until the Exodus led by \textbf{\textit{Moses}}, traditionally dated to the 13th century BCE, after which the Israelites conquered Canaan. \\ \hline

    Short & When did the Duke of Kent die? & Victoria was the daughter of Prince Edward, Duke of Kent and Strathearn, the fourth son of King George III. Both the Duke of Kent and King George III died in \textbf{\textit{1820}}, and Victoria was raised under close supervision by her German-born mother Princess Victoria of Saxe-Coburg-Saalfeld. She inherited the throne aged 18, after her father's three elder brothers had all died, leaving no surviving legitimate children. The United Kingdom was already an established constitutional monarchy, in which the sovereign held relatively little direct political power. Privately, Victoria attempted to influence government policy and ministerial appointments; publicly, she became a national icon who was identified with strict standards of personal morality. \\

    \hline
    
    Short & What is the evaluator called in a breed show? & In conformation shows, also referred to as breed shows, \textbf{\textit{a judge}} familiar with the specific dog breed evaluates individual purebred dogs for conformity with their established breed type as described in the breed standard. As the breed standard only deals with the externally observable qualities of the dog (such as appearance, movement, and temperament), separately tested qualities (such as ability or health) are not part of the judging in conformation shows. \\ 
    \hline

    \end{tabular}%
    \caption{Examples of QA examples with \textit{short} answers where answers are highlighted.}
    \label{appendix:answer_len_examples_2}%

\end{table*}

\subsection{QA examples with \textit{front} answers and \textit{back} answers}

We give some QA examples with character-level answer positions in \textit{front} group and \textit{back} group in Table~\ref{appendix:answer_position_examples_1} and Table~\ref{appendix:answer_position_examples_2}.

\begin{table*}
    \centering

  \begin{tabular}{|p{3cm}|p{4cm}|p{8cm}|}

    \hline
    
    Answer Position & Question & Context \\
    
    \hline
    
     Front &  What are the first names of the men that invented youtube? & According to a story that has often been repeated in the media, \textbf{\textit{Hurley and Chen}} developed the idea for YouTube during the early months of 2005, after they had experienced difficulty sharing videos that had been shot at a dinner party at Chen's apartment in San Francisco. Karim did not attend the party and denied that it had occurred, but Chen commented that the idea that YouTube was founded after a dinner party \"was probably very strengthened by marketing ideas around creating a story that was very digestible\". \\ \hline

    Front & Who became Chairman of the Council of Ministers in 1985? & In the fall of 1985, Gorbachev continued to bring younger and more energetic men into government. On September 27, \textbf{\textit{Nikolai Ryzhkov}} replaced 79-year-old Nikolai Tikhonov as Chairman of the Council of Ministers, effectively the Soviet prime minister, and on October 14, Nikolai Talyzin replaced Nikolai Baibakov as chairman of the State Planning Committee (GOSPLAN). At the next Central Committee meeting on October 15, Tikhonov retired from the Politburo and Talyzin became a candidate. Finally, on December 23, 1985, Gorbachev appointed Yeltsin First Secretary of the Moscow Communist Party replacing Viktor Grishin. \\

    \hline
    
    Front & During what seasons is fog common in Boston? & Fog is fairly common, particularly in \textbf{\textit{spring and early summer}}, and the occasional tropical storm or hurricane can threaten the region, especially in late summer and early autumn. Due to its situation along the North Atlantic, the city often receives sea breezes, especially in the late spring, when water temperatures are still quite cold and temperatures at the coast can be more than 20 °F (11 °C) colder than a few miles inland, sometimes dropping by that amount near midday. Thunderstorms occur from May to September, that are occasionally severe with large hail, damaging winds and heavy downpours. Although downtown Boston has never been struck by a violent tornado, the city itself has experienced many tornado warnings. Damaging storms are more common to areas north, west, and northwest of the city. Boston has a relatively sunny climate for a coastal city at its latitude, averaging over 2,600 hours of sunshine per annum. \\ 
    \hline

    \end{tabular}%
    \caption{Examples of QA examples with answers in \textit{front} group where answers are highlighted.}
    \label{appendix:answer_position_examples_1}%

\end{table*}
\begin{table*}
    \centering

  \begin{tabular}{|p{3cm}|p{4cm}|p{8cm}|}

    \hline
    
    Answer Position & Question & Context \\
    
    \hline
    
    Back & How many murders did Detroit have in 2015? & Detroit has struggled with high crime for decades. Detroit held the title of murder capital between 1985-1987 with a murder rate around 58 per 100,000. Crime has since decreased and, in 2014, the murder rate was 43.4 per 100,000, lower than in St. Louis, Missouri. Although the murder rate increased by 6\% during the first half of 2015, it was surpassed by St Louis and Baltimore which saw much greater spikes in violence. At year-end 2015, Detroit had \textbf{\textit{295}} criminal homicides, down slightly from 299 in 2014. \\
    
    \hline
    
    Back & Who was leading the Conservatives at this time? & Despite being a persistent critic of some of the government's policies, the paper supported Labour in both subsequent elections the party won. For the 2005 general election, The Sun backed Blair and Labour for a third consecutive election win and vowed to give him \"one last chance\" to fulfil his promises, despite berating him for several weaknesses including a failure to control immigration. However, it did speak of its hope that the Conservatives (led by \textbf{\textit{Michael Howard}}) would one day be fit for a return to government. This election (Blair had declared it would be his last as prime minister) resulted in Labour's third successive win but with a much reduced majority. \\
    
    \hline
    
    Back & Who lost the 2015 Nigerian presidential election? & Nigeria is a Federal Republic modelled after the United States, with executive power exercised by the president. It is influenced by the Westminster System model[citation needed] in the composition and management of the upper and lower houses of the bicameral legislature. The president presides as both Head of State and head of the national executive; the leader is elected by popular vote to a maximum of two 4-year terms. In the March 28, 2015 presidential election, General Muhammadu Buhari emerged victorious to become the Federal President of Nigeria, defeating then incumbent \textbf{\textit{Goodluck Jonathan}}. \\
    
    \hline

    \end{tabular}%
    \caption{Examples of QA examples with answers in \textit{back} group where answers are highlighted.}
    \label{appendix:answer_position_examples_2}%

\end{table*}

\subsection{Performance Difference for Text Length and Answer Position Experiments}

We also show the difference of the performance (EM and F-1 score) between QA systems ($QA_{L}-QA_{S}$ and $QA_{F}-QA_{B}$) on subdomains of \textit{text length} and \textit{answer position} in Figure~\ref{fig:text_len_appendix} and Figure~\ref{fig:answer_position_appendix}.

\end{document}